# Multi-modal Wound Classification using Wound Image and Location by Deep Neural Network

D. M. Anisuzzaman[1,#], Yash Patel[1,#], Behrouz Rostami[2], Jeffrey Niezgoda[3], Sandeep Gopalakrishnan[4], and Zeyun Yu[1, 5*]

[1]Department of Computer Science, University of Wisconsin-Milwaukee, Milwaukee, WI, USA;
[2]Department of Electrical Engineering, University of Wisconsin-Milwaukee, Milwaukee, WI 53211, USA
[3]Advancing the Zenith of Healthcare (AZH) Wound and Vascular Center, Milwaukee, WI, USA;
[4]College of Nursing, University of Wisconsin Milwaukee, Milwaukee, WI, USA;
[5]Department of Biomedical Engineering, University of Wisconsin-Milwaukee, Milwaukee, WI, USA.

[*]Corresponding authors:
Zeyun Yu, Professor and Director of Big Data Analytics and Visualization Laboratory, Department of Computer Science, University of Wisconsin-Milwaukee, Milwaukee, WI, USA. Email: yuz@uwm.edu

[#] Authors have an equal contribution.

## Abstract

Wound classification is an essential step of wound diagnosis. An efficient classifier can assist wound specialists in classifying wound types with less financial and time costs and help them decide an optimal treatment procedure. This study developed a deep neural network-based multi-modal classifier using wound images and their corresponding locations to categorize wound images into multiple classes, including diabetic, pressure, surgical, and venous ulcers. A body map is also developed to prepare the location data, which can help wound specialists tag wound locations more efficiently. Three datasets containing images and their corresponding location information are designed with the help of wound specialists. The multi-modal network is developed by concatenating the image-based and location-based classifier's outputs with some other modifications. The maximum accuracy on mixed-class classifications (containing background and normal skin) varies from 77.33% to 100% on different experiments. The maximum accuracy on wound-class classifications (containing only diabetic, pressure, surgical, and venous) varies from 72.95% to 98.08% on different experiments. The proposed multi-modal network also shows a significant improvement in results from the previous works of literature.

## I. Introduction

More than 8 million people are suffering from wounds, and the medicare cost related to wound treatments ranged from $28.1 billion to $96.8 billion, according to a 2018 retrospective analysis [1]. This immense number can give us an idea of the population related to wound and their care and management. The most common types of wounds/ulcers are diabetic foot ulcer (DFU), venous leg ulcer (VLU), pressure ulcer (PU), and surgical wound (SW). About 34% of people with diabetes have a lifetime risk of developing a DFU, and more than 50% of diabetic foot ulcers become infected [2]. About 0.15% to 0.3% of people are suffering from active VLU worldwide [3]. A pressure ulcer is another significant wound, and 2.5 million people are affected each year [4]. Yearly about 4.5% of people have a surgery that leads to a surgical wound [5].

The above statistics show that wounds have caused a huge financial burden and may even be life-threatening to patients. An essential part of wound care is to differentiate among different types of wounds (DFU, VLU, PU, SW, etc.) or among wound conditions (infection vs. non-infection, ischemia vs. non-ischemic, etc.). To prepare proper medication and treatment guidelines, physicians must first detect the correct class of the wound. Until the recent advancement of artificial intelligence (AI), wounds were manually classified by wound specialists. AI can save both time and cost and, in some cases, may give better predictions than humans. In recent years, AI algorithms have evolved into so-called data-driven techniques without human or expert intervention, as compared to the early generations of AI that were rule-based, relying mainly on an expert's knowledge [6]. This research is focused on wound type classification by using a data-driven AI technique named Deep Learning (DL).

Deep learning is prevalent in image processing, with a huge success in medical image analysis. In the general field of image processing and study, some widely used DL algorithms are Convolutional Neural Networks (CNN), Deep Belief Networks (DBN), Deep Boltzmann Machines (DBM), and Stacked (Denoising) Autoencoders [7]. In addition, some of the most common DL methods for medical image analysis include LeNet, AlexNet, VGG 19, GoogleNet, ResNet, FCNN, RNNs, Auto-encoders, and Stacked Auto-encoders, Restricted Boltzmann Machines (RBM), Variational Auto-encoders and Generative Adversarial Networks [8]. Bakator et al. [9] reviewed CNN, RBM, Self-Advised Support Vector Machine (SA-SVM), Convolutional Recurrent Neural Network (CRNN), DBN, Stacked Denoising Autoencoders (SDAE), Undirected Graph Recursive Neural Networks (UGRNN), U-NET, and Class Structure-Based Deep Convolutional Neural Network (CSDCNN) as deep learning methods in the field of medical diagnosis.

Though there exists some feature-based machine learning and end-to-end deep learning models for image-based wound classification, the classification accuracy is limited due to incomplete information considered in the classifiers. The novelty of the present research is to add wound location as a vital feature to obtain a more accurate classification result. Wound location is a standard entry for electronic health record (EHR) document, which many wound physicians utilize for wound diagnosis and prognosis. Unfortunately, these locations are documented manually without any specific guideline, which leads to some inconsistency. In the current work, we developed a body map from which one can select the location of the wound visually and accurately. Then, for each wound image, the wound location is set through the body map, and the location is indexed according to the image file name. Finally, the developed classifier is trained with both image (gained through convolution) and location features and produces superior classification performance compared to image-based wound classifiers. A basic workflow of this research is shown in Figure 1. The developed wound classifier takes both wound image and location as inputs and outputs the corresponding wound class.

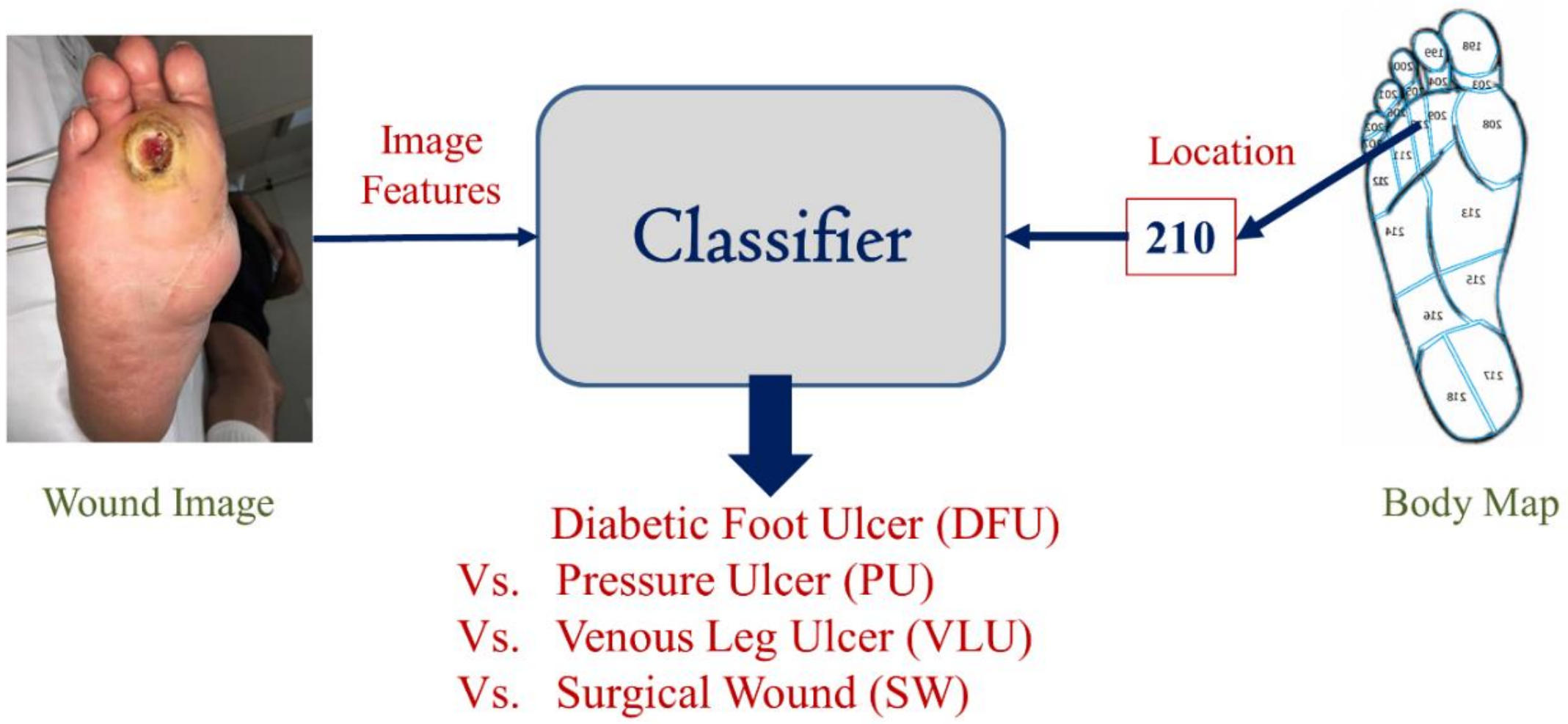

**Figure 1:** Workflow of this research

The remainder of the work is organized as follows. Related works on wound classification are discussed in Section 2. Section 3 discusses the methodology, where dataset, body map, and classification models are described. In Section 4, experimental setup, results and comparison, and discussion on the results are presented. Finally, the paper is concluded, and some remarks on future directions are given.

## II. Related Work

Wound classification includes wound type classification, wound tissue classification, burn depth classification, etc. Wound type classification considers different types of wounds and non-wounds (normal skin, background, etc.). Background versus DFU, normal skin versus PU, DFU versus PU are examples of binary wound type classification, where DFU versus PU versus VLU is an example of multi-class wound type classification. Wound tissue classification differentiates among different types of tissues (granulation, slough, necrosis, etc.) within a specific wound. Burn depth classification measures the depth (superficial dermal, deep dermal, full-thickness, etc.) of the burn wound. As this research focuses on wound type classification, this section discusses existing data-driven wound type classification works. We presented wound type classification methods into two sections: machine learning-based and deep learning-based methods.

### A. Machine Learning-based Methods

A machine learning approach was proposed by Abubakar et al. [10] to differentiate burn wounds and pressure ulcers. Features were extracted using pre-trained deep architectures like VGG-face, ResNet101, and ResNet152 from the images and then fed into an SVM classifier to classify the images into burn or pressure wound classes. The dataset used in this study includes 29 pressure and 31 burn wound images obtained from the internet and a hospital, respectively. After augmentation, they had three categories: burn, pressure, and healthy skin, with 990 sample images in each class. Several experiments, including binary classification (burn or pressure) and 3-class classification (burn, pressure, and healthy skin), were conducted.

Goyal et al. [11] used traditional machine learning, deep learning, and ensemble CNN models for binary classification of ischemia versus non-ischemia and infection versus non-infection on DFU images. The authors developed a dataset containing 1459 DFU images that two healthcare professionals labeled. For traditional machine learning, the authors used BayesNet, Random Forest, and Multilayer perceptron. Three CNN networks (InceptionV3, ResNet50, and InceptionResNetV2) were used as deep-learning approaches. The ensemble CNN contains an SVM classifier which takes the bottleneck features of three CNN networks as input. The test evaluation shows that traditional machine learning methods perform the worst, followed by deep-learning networks, while the ensemble CNN performs the best in both binary classifications. The authors reported an accuracy of 90% for ischemia classification and 73% for infection classification.

### B. Deep Learning-based Methods

There are some deep learning-based wound type classification works. One of the advantages of deep learning models is that they do not require any manually crafted features as input, but instead, they find the features with the given input and output. For example, a novel CNN architecture named DFUNet was developed by Goyal et al. [12] for binary classification of healthy skin and DFU skin. A dataset of 397 wound images was presented, and data augmentation techniques were applied to increase the number of images. The proposed DFUNet utilized the idea of concatenating the outputs of three parallel convolutional layers with different filter sizes. An accuracy of 92.5% was reported for the proposed method.

A CNN-based method is proposed by Aguirre et al. [13] for VLU versus non-VLU classification from ulcer images. In this study, a pre-trained VGG-19 network was used for classifying the ulcer images in the mentioned two categories. First, a dataset of 300 pictures annotated by a wound specialist was proposed, and data pre-processing and augmentation were conducted before the network training. Then, the VGG-19 network was pre-trained using another dataset of dermoscopic images. The authors reported 85%, 82%, and 75% accuracy, precision, and recall.

Shenoy et al. [14] proposed a CNN-based method for binary classification of wound images. In this study, they used a dataset of 1335 wound images collected via smartphones and the internet. The authors considered nine different labels (wound, infection (SSI), granulation tissue, fibrinous exudates, open wound, drainage, steri strips, staples, and sutures) for the dataset, where for each label, two subcategories (positive and negative) were considered. The authors used a modified VGG16 network named WoundNet as the classifier, pre-trained using the ImageNet dataset. In addition, the researchers created another network called Deepwound, an ensemble model that averages the results of three individual models. The reported accuracy varies from 72% (drainage) to 97% (steri strips), where the accuracy for the class “wound” is 82%.

A binary patch classification of normal skin versus abnormal skin (DFU) is performed by Alzubaidi et al. [15] with a novel deep convolutional neural network named DFU_QUTNet. First, the authors introduced a new dataset consisting of 754-foot images from a diabetic hospital center in Iraq. From these 754 images, 542 normal skin patches and 1067 DFU patches were generated.

Then, in the augmentation step, they multiplied the number of training samples by 13, using flipping, rotating, and scaling transformations. The proposed network is a deep architecture with 58 layers, including 17 convolutional layers. The performance of their proposed method was compared with those of other deep CNNs like GoogLeNet, VGG16, and AlexNet. The maximum reported F1-Score was 94.5% obtained from combining the DFU_QUTNet architecture with SVM.

Rostami et al. [16] proposed an end-to-end ensemble DCNN-based classifier to classify the entire wound images into multiple classes, including surgical, diabetic, and venous ulcers. The output classification scores of two classifiers based on patch-wise and image-wise strategies are fed into a Multi-Layer Perceptron to provide a superior classifier. A new dataset of authentic wound images containing 538 images from four different types of wounds was introduced in this research. The reported maximum and average classification accuracy values are 96.4% and 94.28% for binary and 91.9% and 87.7% for 3-class classification.

Sarp et al. [17] classified chronic wounds into four classes (diabetic, lymphovascular, pressure injury, and surgical) by using an explainable artificial intelligence (XIA) approach to provide transparency on the neural network. The dataset contains 8690 wound images collected from the data repository of eKare, Inc. Mirroring, rotation, and horizontal flip augmentations were used to increase the number of wound images and to balance the number of pictures in each class. Transfer learning on the VGG16 network was used as the classifier model. The authors reported an average F1-score of 0.76 as the test result. The XIA technique can provide explanation and transparency for the wound image classifier and why the model would think a particular class may be present.

Though some wound type classification works from wound images exist, to the best of our knowledge, there is no automated wound classification work based on the wound location feature. This research is the first work that incorporates wound location for automatic wound type classification and proposes a multi-modal network that uses both wound image features and location features to classify a wound.

# III. Methodology

## A. Dataset

In this research, three different datasets are used for the robustness and reliability of the model performance. A brief discussion of these datasets is given below:

### 1. AZH Dataset

AZH dataset is collected over a two-year clinical period at the AZH Wound and Vascular Center in Milwaukee, Wisconsin. The dataset includes 730 wound images in .jpg format. The images are of various sizes, where the width ranging from 320 to 700 pixels and the height ranging from 240 to 525 pixels. These images contain four different wound types: venous, diabetic, pressure, and surgical. iPad Pro (software version 13.4.1) and a Canon SX 620 HS digital camera are used to capture the images, and labeling is done by a wound specialist from the AZH Wound and Vascular Center. For most images in our dataset, each image is taken from a separate patient.

But there are a few cases where multiple photos were taken from the same patient at different body sites or various healing stages. For the latter case, the wound shapes are different, so they are considered separate images.

### 2. Medetec Dataset

Medetec wound database [18] contains free stock images of all types of open wounds. We randomly collect 358 images from these three categories: diabetic, pressure, and arterial and venous leg ulcer. The arterial and venous leg ulcer images are not separated in the Medetec database, so we consider them the same category. This dataset does not contain any surgical wound images. All the images are in .jpg format, where the weight varies from 358 to 560 pixels, and the height varies from 371 to 560 pixels.

### 3. AZHMT Dataset

This dataset is the mixer of all the images from the AZH and Medetec datasets. This dataset contains 1088 wound images in .jpg format. AZHMT includes four wound classes: diabetic, pressure, surgical, and arterial + venous leg ulcers. The width of these images varies from 320 to 700 pixels, and the height ranging from 240 to 560 pixels. AZHMT dataset is created for the reliability and robustness testing of the developed model.

## B. Body Map for Location

A body map is a labeled, simplified, and symbolic diagram of the entire body of the person, which should be phenotypically right [19]. Medical practitioners use body maps to locate bruises, wounds, or body breakage on a patient's body. Moreover, forensic scientists use body diagrams to help them identify and determine body changes during a postmortem examination. Doctors use body maps to analyze the location of a given infection on patients [20]. A detailed body map helps doctors determine which other part of the body to be cautious about during the wound's rehabilitation process. Moreover, a body map is a piece of medical evidence during a scientific study. A health practitioner can use notable body changes shown by a body map as a backup of an existing ailment affecting the patient internally.

Wound history is another benefit attributed to efficient body mapping. A doctor can collect information on the wound's cause, previous measures adopted in providing care to the wound, and underlying health complications such as diabetes that would deter the healing process. Detailed wound history needs to be collected and all causes explored to avoid delayed or static healing. Body mapping contributes to wound treatment localization significantly. Pain location, activities of daily living, and the type of wound are factors that a doctor should consider in the localization process. Wilson asserts that a wound located on the heel area and a wound located on the lower abdomen or joint area would not have a similar rehabilitation technique. The wound situated on the heel would need the doctor to consider the weight issue instead of the wound located on the lower abdomen. Therefore, the doctor would need to localize their examination and the treatment process depending on the wound's location and other external factors that directly affect the wound weight and joint movement [20].

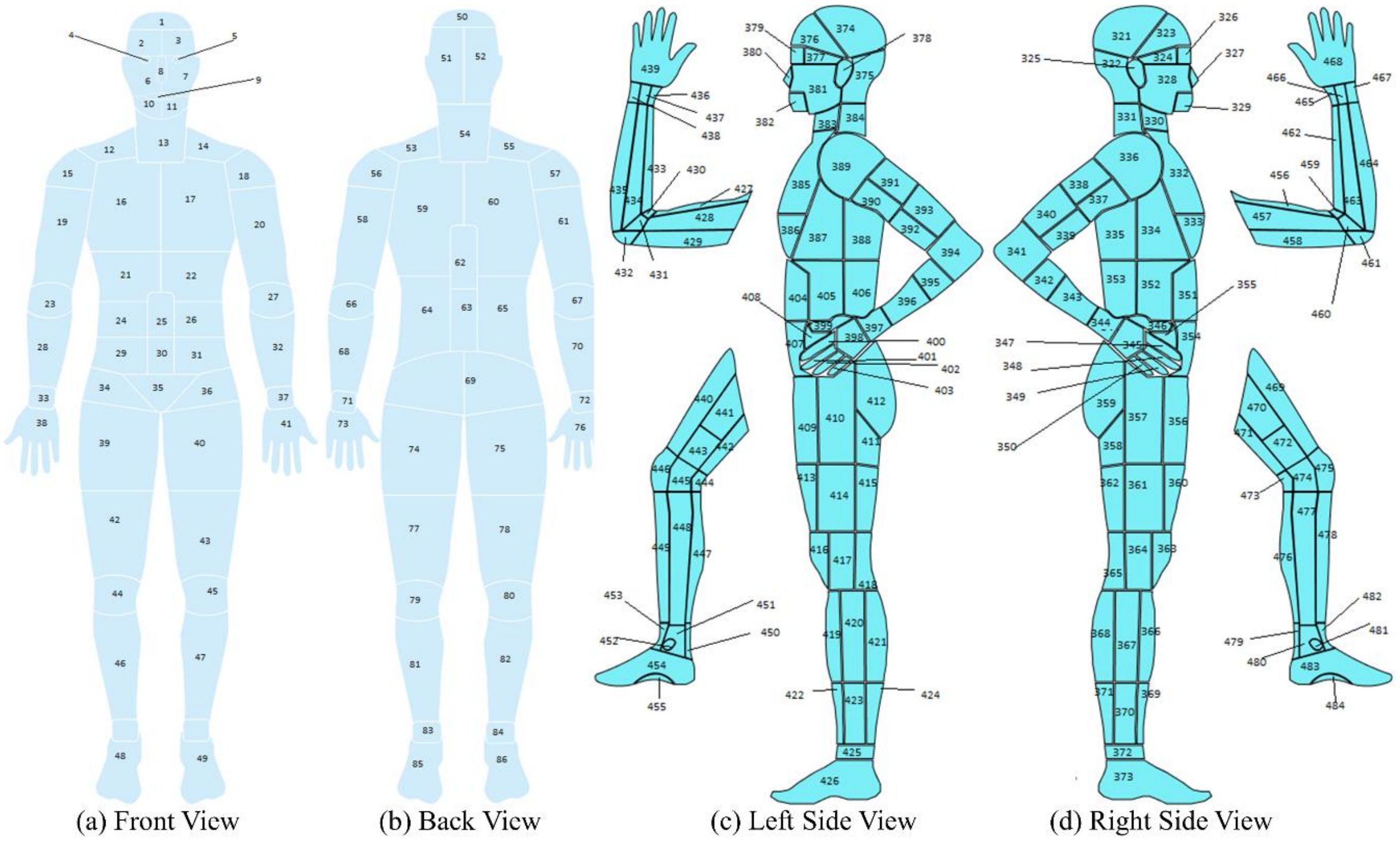

(i) Full Body View

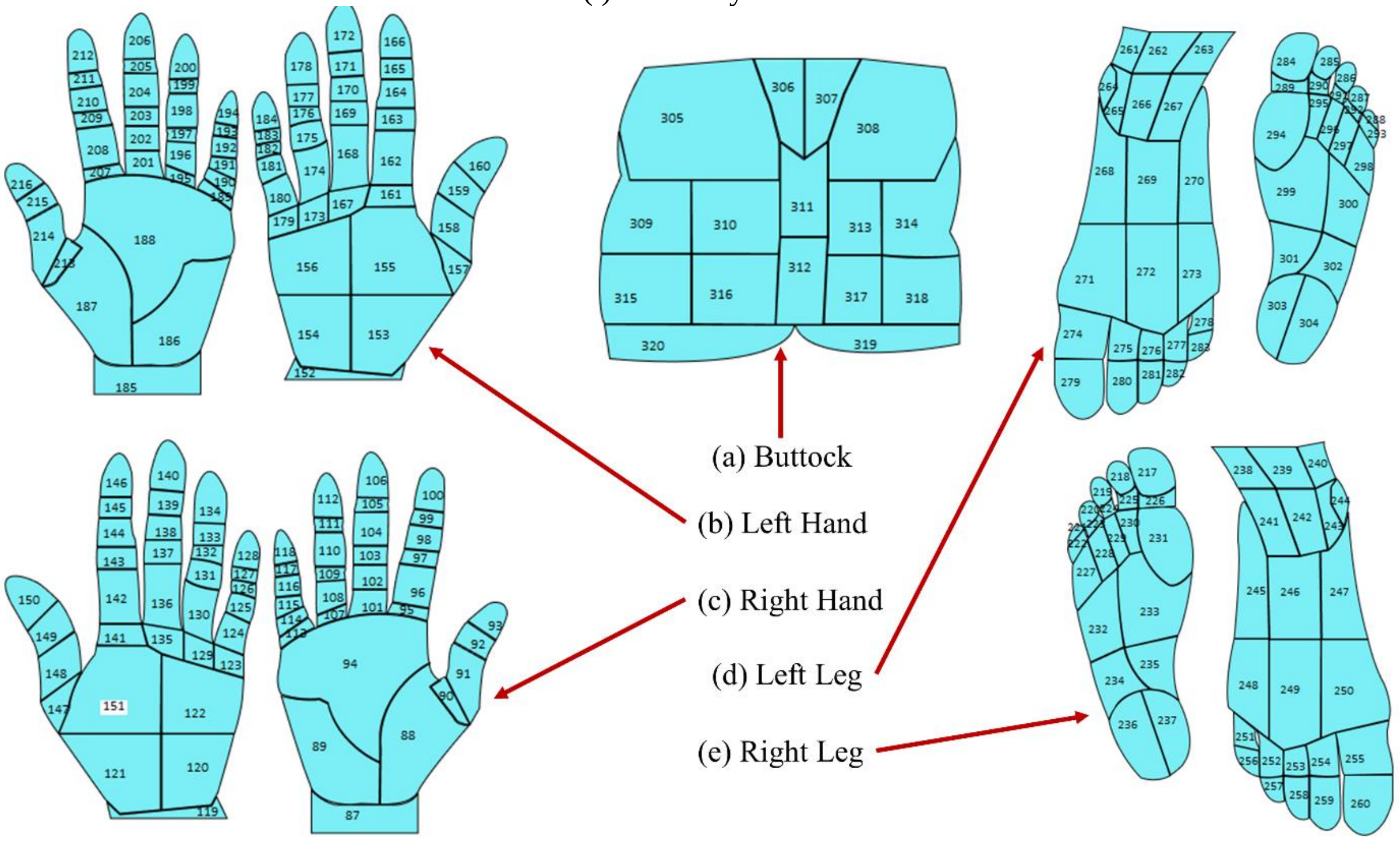

(ii) Detailed Body View

**Figure 2:** Body Map for Location Selection

A simplified body map with 484 total parts is designed to avoid the body map's complexity. The body map is prepared using PaintCode [21]. The initial reference to the body map is obtained from [22]–[24]. The ground truth diagram for the design is based on Original Anatomy Mapper [25]. Each label and outline are directly paired to the labeling provided by the anatomy mapper [25]. To avoid extreme complexity of drawing every detailed feature of the body map, a total of 436 feature or region is pre-selected and approved by wound professionals at AZH wound and vascular center. The developed body map is shown in Figure 2. Here each number represents a location. Few examples of the locations and their corresponding numbers are shown in Table 1.

**Table 1:** Examples of Locations and their Corresponding Mapping

| **Left Hand Front** | | **Right Leg Bottom** | | **Buttock** | |
|---|---|---|---|---|---|
| **Location** | **Reference Number** | **Location** | **Reference Number** | **Location** | **Reference Number** |
| Left Dorsal Wrist | 152 | Right Distal Plantar First Toe | 217 | Left Posterior Lower Back | 305 |
| Left Proximal Lateral Dorsal Hand | 153 | Right Proximal Plantar First Toe | 226 | Superior Gluteal | 311 |
| Left Proximal Medial Dorsal Hand | 154 | Right Distal Lateral Mid Plantar Foot | 232 | Inferior Gluteal | 312 |
| Left Distal Phlanax of Dorsal Little Finger | 184 | Right Medial Heel | 237 | Left Gluteal Fold | 320 |

## C. Dataset Processing

All datasets go through three significant steps: region of interest (ROI) cropping, location labeling, and data augmentation. The ROI of a wound image means the wound and some of its surrounding area (healthy skin) that contains the essential information of a wound. From each image, single or multiple ROIs have cropped automatically by using our developed wound localizer [26]. The extracted ROIs are rectangular, but their height and weight are different depending on the wound size. Then all the ROI's locations are labeled by a wound specialist of the AZH wound and vascular center. The location labeling is done by using our developed body map. As our body map represents each location with a unique number, each ROI is tagged with a location number for model training. Finally, rotation and flipping augmentations are used for each ROI to increase the data numbers. Total five augmentations are applied to each ROI: horizontal and vertical flip, 25-degree, 45-degree, and 90-degree rotations. As wound location does not change with image augmentation, the location number is repeated for each augmented image. The dataset processing step is illustrated in Figure 3.

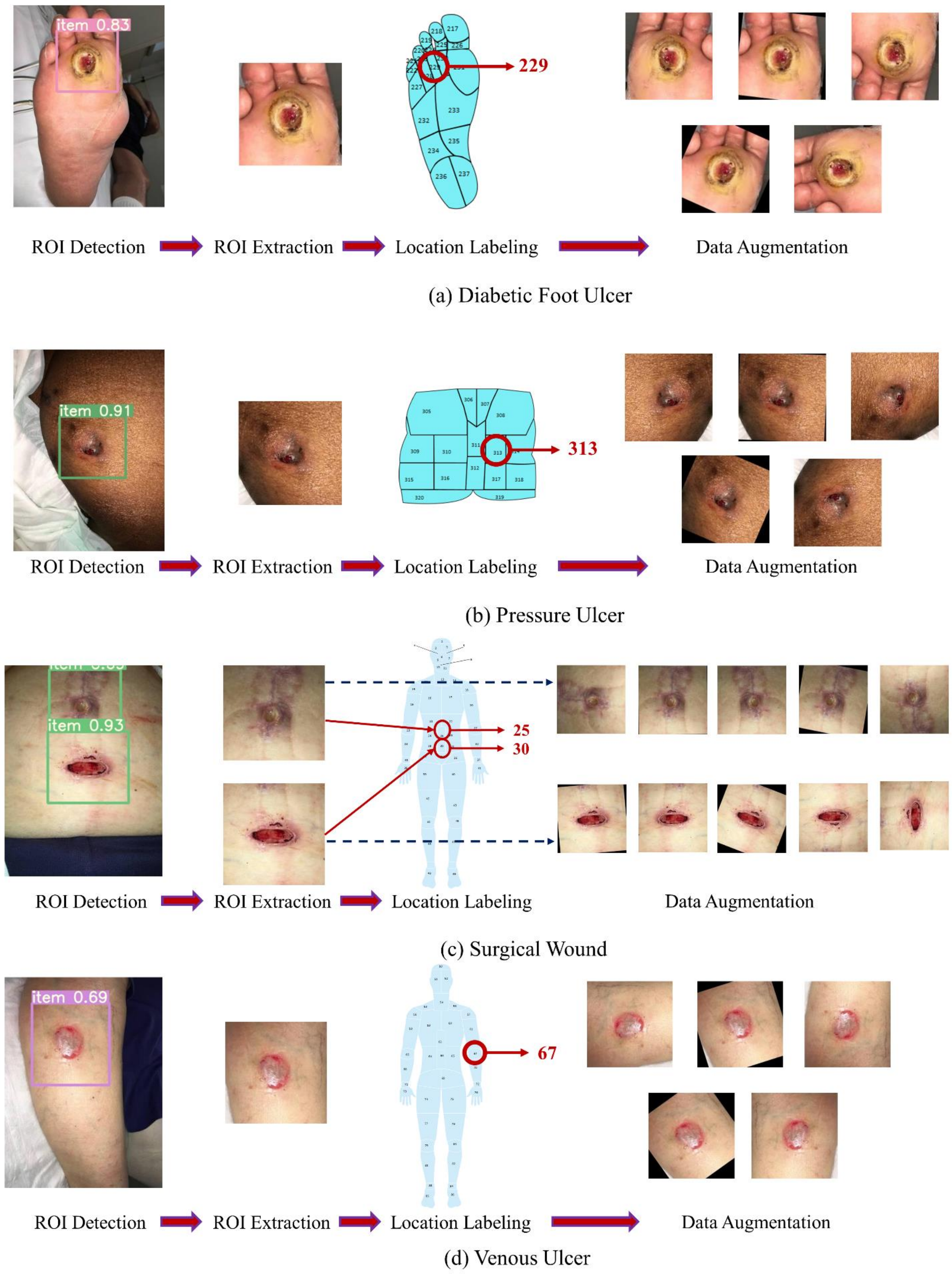

**Figure 3:** Dataset Processing

Each dataset (ROI) is divided into 60% training, 15% validation, and 25% test sets. First, the 25% test set is created from a random selection of the wound images to ensure no overlap between training and test sets. The validation set is also created randomly during the time of training. Next, the 75% training and validation datasets are augmented, while test images do not go through data augmentation. Two non-wound classes named normal skin and background are created by manually cropping corresponding ROIs from the original images. A wound specialist does the location tagging for healthy skin. As the background ROIs do not represent any location of our developed body map, each ROI is tagged with a location number '-1'. Table 2 shows the number of images of all three datasets. All the six classes, diabetic, venous, arterial + venous, pressure, surgical, background, and normal skin, are represented with the following abbreviations D, V, A+V, P, S, BG, and N, respectively.

**Table 2:** Description of all Datasets

| Dataset ➡ | AZH | | | Medetec | | | AZHMT | | |
|---|---|---|---|---|---|---|---|---|---|
| **Class** | **Training + Validation** | **Test** | **Total** | **Training + Validation** | **Test** | **Total** | **Training + Validation** | **Test** | **Total** |
| **Background (BG)** | 450 | 25 | 475 | 0 | 0 | 0 | 450 | 25 | 475 |
| **Normal Skin (N)** | 450 | 25 | 475 | 0 | 0 | 0 | 450 | 25 | 475 |
| **Diabetic (D)** | 834 | 46 | 880 | 330 | 19 | 349 | 1164 | 65 | 1229 |
| **Pressure (P)** | 600 | 34 | 634 | 822 | 46 | 868 | 1422 | 80 | 1502 |
| **Surgical (S)** | 732 | 42 | 774 | 0 | 0 | 0 | 732 | 42 | 774 |
| **Venous (V)** | 1110 | 62 | 1172 | 0 | 0 | 0 | 0 | 0 | 0 |
| **Arterial + Venous (A+V)** | 0 | 0 | 0 | 456 | 25 | 481 | 1566 | 87 | 1653 |
| **Total** | **4176** | **234** | **4410** | **1608** | **90** | **1698** | **5784** | **324** | **6108** |

## D. Model

We see that our dataset contains both image and categorical (wound location) data from the above discussion. We use Keras Functional API [27] to develop a network that can handle multiple inputs and mixed data. The Functional API is more flexible than the Sequential API, which can control models with non-linear topology, shared layers, and even multiple inputs or outputs. Considering a deep learning model as a directed acyclic graph (DAG) of layers, the functional API is a way to build graphs of layers.

Figure 4 shows the architecture of our wound-type classification network. Two separate neural networks for each data type are used to work with both image and location data. These networks are then considered input branches, and their outputs are combined into a final neural network. We address the image network as Wound Image Classifier (WIC) network, the location network as Wound Location Classifier (WLC) network, and the combined network as Wound Multimodality Classifier (WMC) network. The output of this WMC network is the probability of the wound class.

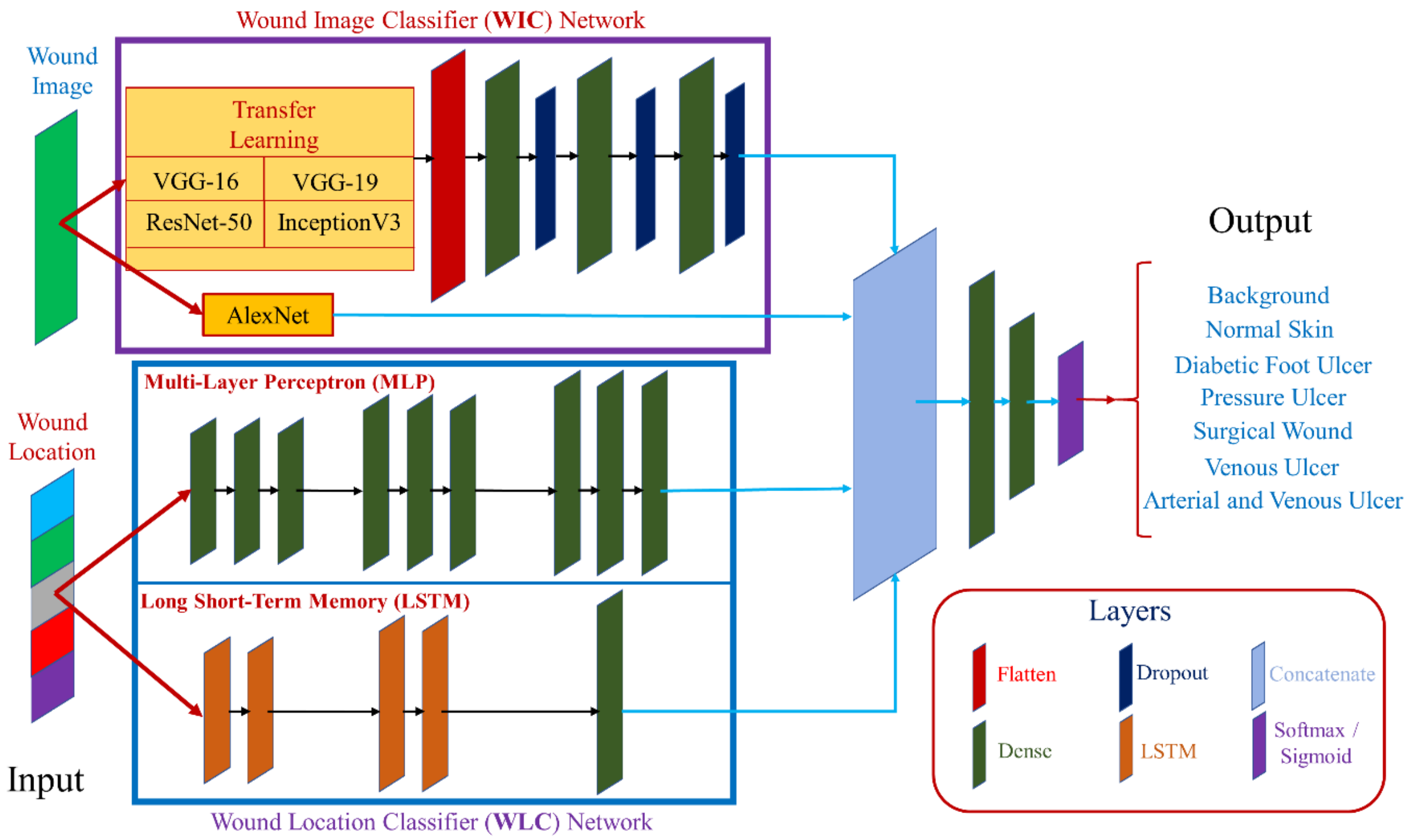

**Figure 4:** Wound Multimodality Classifier (WMC) Network Architecture

It is imperative for the multi-modal network (WMC) to arrange the data in the correct order. The output for the image and location data must be consistent, so the final combined (WMC) neural network must be fed with the right ordered data simultaneously. For example, to train the WMC network properly, we must give the output of the WIC network for the 148$^{th}$ DFU image and the output of the WLC network for the 148$^{th}$ DFU wound's location as the input at the same time to the WMC network. If the data are not ordered correctly, the WMC network may have the WIC network's output for the 148$^{th}$ DFU image and the WLC network's output for the 55$^{th}$ PU wound's location as input at the same time, which will lead to a wrong classification. This arrangement is taken care of by giving each ROI a unique index number and tagging the corresponding location to that index number.

### D.1 Wound Image Classifier (WIC) Network

The wound image classifier (WIC) network is built upon transfer learning, except the AlexNet [28]. Transfer learning means taking advantage of features learned on one problem and using them in another similar situation. This method is proper when the dataset in hand is little in number to train a full-scale model from scratch, and the memory power is limited to train a vast deep learning model. The most commonly used workflow of transfer learning is: 1) take a previously trained model's layers, 2) freeze the layers, 3) add some new, trainable layers on top of the frozen layers, which will learn to turn the old features into predictions on a new dataset, and 4) train the new layers on the new dataset [29]. There are 26 deep learning models in Keras Applications [30], among which we choose four top-rated classification models: VGG16 [31], VGG19 [32], ResNet50 [33], and InceptionV3 [34]; and take their previously trained layers to apply transfer learning. All the layers, except the top layer, are frozen for all these four models, and three Dense layers with dropout layers are added (Figure 4, top WIC box) for training on our wound datasets. All three Dense layers contain 512 trainable neurons, with all having the ReLU activation. The AlexNet [28] is implemented following the original architecture. The output layer is added with either softmax or sigmoid layer for multi-class or binary-class classification for all the models, respectively.

#### D.2 Wound Location Classifier (WLC) Network

The Wound Location Classifier (WLC) network can classify wound locations using either a Multi-Layer Perceptron (MLP) or Long Short-Term Memory (LSTM) network. As the location data is categorical, we used one-hot encoding to represent the data, representing each input to the WLC network as a one-hot vector. The WLC network handles only one categorical data (location), for which the architecture of the network is kept simple. With a deeper network, the accuracy does not improve (sometimes decreases), and resources (time and memory) becomes expensive. The MLP network contains nine Dense layers, with all having the ReLU activation. The first three layers contain 128 neurons, the following three layers contain 256 neurons, and the last three layers contain 512 neurons (Figure 4, middle MLP box). The LSTM contains four LSTM layers, followed by a Dense layer, with all having the ReLU activation. The first two layers contain 32 neurons, followed by two LSTM layers having 64 neurons each, and finally, the Dense layer contains 512 neurons (Figure 4, bottom LSTM box). The output layer is added with either softmax or sigmoid layer for multi-class or binary-class classification for all the models, respectively.

#### D.3 Wound Multimodality Classifier (WMC) Network

As discussed earlier, the Wound Multimodality Classifier (WMC) network is designed using Keras Functional API [27], which can predict the wound classes based on both wound image and location information. At first, the image data goes through the WIC network, and the location data goes through the WLC network, and the outputs of the networks are concatenated. Then, two Dense layers are added after concatenation to learn from the merged features. These Dense layers contain 512 and 256 neurons, respectively. Finally, the output layer is added with either a softmax or sigmoid layer for multi-class or binary-class classification.

## IV. Experiment & Result & Discussion

### A. Experimental Setup

Lots of experiments are performed with different setups. Classification between D vs. V, D vs. S, N vs. D, etc. are some examples of binary classification, and D vs. P vs. S, BG vs. N vs. S vs. V, BG vs. N vs. D vs. P vs. S vs. V, etc. are some examples of multi-class classification. In the WMC network, all combinations of the WIC and WLC networks (AlexNet+MLP, AlexNet+LSTM, ResNet50+MLP, VGG16+LSTM, etc.) are applied for the four wound class classification (D vs. P vs. S vs. V) on the AZH dataset. Based on the results (discussed later), the best two combinations are applied for the other multi-modal classifications.

All the models are written in Python programming language using the Keras deep learning framework and trained on an Nvidia GeForce RTX 2080Ti GPU platform. All models are trained for 250 epochs with a batch size of 25, a learning rate of 0.001, and an Adam optimizer. Two callbacks are used with the best validation accuracy and the best combination of validation and training accuracy saving. For multi-class classification and binary class classification, *sparse_categorical_crossentropy* and *binary_crossentropy* loss functions are used, respectively.

To investigate the classification performance, we use accuracy as the performance metric. Accuracy is the ratio of correctly predicted data to the total amount of data. To evaluate binary classifications, we use precision, recall, and f1-score as performance metrics as well. Equations 1 to Equation 4 show the related formulae for these evaluation metrics. In these equations, TP, TN, FP, and FN, represent True Positive, True Negative, False Positive, and False Negative measures. More details about these equations can be found in [35].

$$Accuracy = \frac{TP + TN}{TP + FP + FN + TN} \tag{1}$$

$$Precision = \frac{TP}{TP + FP} \tag{2}$$

$$Recall = \frac{TP}{TP + FN} \tag{3}$$

$$F1 - Score = 2 \times \frac{Recall \times Precision}{Recall + Precision} \tag{4}$$

## B. Results

### B.1 Selecting Best Combination

Four wound class classification (D vs. P vs. S vs. V) on the AZH dataset is chosen to select the best combinations for the WMC network. This classification is the most challenging classification task, as there are no normal skin (N) or background (BG) images in the experiment. Table 3 shows the results of this experiment. We also present the results on the original dataset (without any augmentation) for this experiment to show the effect (improvement) of data augmentation. The performances of MLP and LSTM are similar on the WLC network, and the VGG16 and VGG19 perform best on the WIC network. Their combinations: VGG16+MLP, VGG19+MLP, VGG16+LSTM, and VGG19+LSTM, also work best for the WMC network. The performance of AlexNet+MLP, AlexNet+LSTM, ResNet50+MLP, and ResNet50+LSTM are very poor. The InceptionV3+MLP and InceptionV3+LSTM performances are also not good enough to apply for all the experiments. Running all these combinations for many experiments is also expensive (both with time and memory). So, from these results, we apply InceptionV3+MLP and InceptionV3+LSTM combinations along with the best four combinations (VGG16+MLP, VGG19+MLP, VGG16+LSTM, and VGG19+LSTM) for 5-class classifications and 6-class classification. With not getting good results from the InceptionV3+MLP and InceptionV3+LSTM combinations, we discarded them from applying on 4-class, 3-class, and binary classifications on the WMC network.

**Table 3:** Four Wound Class Classification (D vs. P vs. S vs. V) on AZH Dataset

| Input | Model | Original Dataset | Augmented Dataset |
|---|---|---|---|
| | | Accuracy | Accuracy |
| Location | MLP | 66.30% | 71.74% |
| | LSTM | 66.85% | 72.28% |
| Image | AlexNet | 35.33% | 37.50% |
| | VGG16 | 65.76% | 71.73% |
| | VGG19 | 56.52% | 63.04% |
| | InceptionV3 | 51.09% | 56.52% |
| | ResNet50 | 33.70% | 33.70% |
| Image + Location | AlexNet+MLP | 55.43% | 61.41% |
| | VGG16+MLP | 77.17% | 78.26% |
| | VGG19+MLP | 62.50% | 72.28% |
| | InceptionV3+MLP | 61.41% | 70.11% |
| | ResNet50+MLP | 63.04% | 66.85% |
| | AlexNet+LSTM | 58.15% | 66.85% |
| | VGG16+LSTM | 72.83% | **79.35%** |
| | VGG19+LSTM | 71.20% | 76.63% |
| | InceptionV3+LSTM | 64.67% | 69.02% |
| | ResNet50+LSTM | 33.70% | 34.79% |

### B.2 Experiment on AZH Dataset

A classification between all the classes is performed on the AZH dataset. Table 4 shows the results of this six-class classification (BG vs. N vs. D vs. P vs. S vs. V). We achieve the highest accuracy of 82.91% with the multi-modal (WMC) network using the VGG19+MLP combination, where the highest accuracies reach from WLC and WIC networks are 60.68% and 75.64% using MLP and VGG16 networks, respectively.

**Table 4:** Six-Class Classification (BG vs. N vs. D vs. P vs. S vs. V) on AZH Dataset

| Input | Model | Accuracy |
|---|---|---|
| Location | MLP | 60.68% |
| | LSTM | 60.26% |
| Image | VGG16 | 75.64% |
| | VGG19 | 64.96% |
| | InceptionV3 | 54.70% |
| Image + Location | VGG16+MLP | 79.49% |
| | VGG19+MLP | **82.91%** |
| | InceptionV3+MLP | 65.38% |
| | VGG16+LSTM | 76.92% |
| | VGG19+LSTM | 69.68% |
| | InceptionV3+LSTM | 64.10% |

Four five-class classifications are performed on the AZH dataset. The classifications are 1) BG vs. N vs. D vs. P vs. V, 2) BG vs. N vs. D vs. S vs. V, 3) BG vs. N vs. D vs. P vs. S, and 4) BG vs. N vs. P vs. S vs. V. We achieve the highest accuracy of 84.38%, 90.00%, 77.33%, and 80.32% for classification number 1), 2), 3), and 4), respectively. In all four classifications, the highest accuracy is achieved with the multi-modal (WMC) networks. Table 5 shows the detailed results of these classifications.

**Table 5:** Four Five-Class Classifications on AZH Dataset

| Classifications | | BG-N-D-P-V | BG-N-D-S-V | BG-N-D-P-S | BG-N-P-S-V |
|---|---|---|---|---|---|
| **Input** | **Model** | **Accuracy** | **Accuracy** | **Accuracy** | **Accuracy** |
| Location | MLP | 65.10% | 68.00% | 58.14% | 70.21% |
| | LSTM | 66.15% | 67.50% | 58.72% | 72.34% |
| Image | VGG16 | 69.79% | 70.50% | 64.53% | 75.53% |
| | VGG19 | 76.56% | 74.50% | 67.44% | 72.34% |
| | InceptionV3 | 52.60% | 57.00% | 47.09% | 52.66% |
| Image + Location | VGG16+MLP | 77.08% | 85.00% | **77.33%** | 75.53% |
| | VGG19+MLP | **84.38%** | **90.00%** | 75.58% | 76.06% |
| | InceptionV3+MLP | 56.77% | 72.00% | 62.79% | 56.38% |
| | VGG16+LSTM | 82.29% | 82.00% | 70.35% | **80.32%** |
| | VGG19+LSTM | 83.85% | 77.50% | 68.02% | 74.47% |
| | InceptionV3+LSTM | 59.90% | 58.00% | 58.72% | 37.23% |

Six four-class classifications are performed on the AZH dataset, along with one wound class classification (shown in Table 3). The classifications are: 1) BG vs. N vs. D vs. V, 2) BG vs. N vs. P vs. V, 3) BG vs. N vs. S vs. V, 4) BG vs. N vs. D vs. P, 5) BG vs. N vs. D vs. S, and 6) BG vs. N vs. P vs. S. We achieve the highest accuracy of 95.57%, 94.52%, 94.81%, 86.15%, 89.86%, and 88.89% for classification number 1), 2), 3), 4), 5), and 6), respectively. In addition, in all six classifications, the highest accuracy is achieved with the multi-modal (WMC) networks. Table 6 shows the detailed results of these classifications.

**Table 6:** Six Four-Class Classifications on AZH Dataset

| Classifications | | BG-N-D-V | BG-N-P-V | BG-N-S-V | BG-N-D-P | BG-N-D-S | BG-N-P-S |
|---|---|---|---|---|---|---|---|
| **Input** | **Model** | **Accuracy** | **Accuracy** | **Accuracy** | **Accuracy** | **Accuracy** | **Accuracy** |
| Location | MLP | 72.78% | 72.60% | 63.64% | 63.08% | 75.36% | 63.49% |
| | LSTM | 74.05% | 52.74% | 74.68% | 63.85% | 74.64% | 69.84% |
| Image | VGG16 | 93.67% | 89.73% | 87.66% | 82.31% | 77.54% | 83.33% |
| | VGG19 | 89.87% | 86.99% | 88.31% | 80.00% | 81.88% | 83.33% |
| Image + Location | VGG16+MLP | **95.57%** | 91.78% | **94.81%** | 81.54% | 87.68% | 85.71% |
| | VGG19+MLP | 94.30% | 91.10% | 91.56% | **86.15%** | **89.86%** | 84.13% |
| | VGG16+LSTM | 92.41% | **94.52%** | 92.21% | 83.85% | 83.33% | 87.30% |
| | VGG19+LSTM | 91.77% | 89.73% | 92.86% | 80.00% | 81.16% | **88.89%** |

Four three-wound-class classifications are performed on the AZH dataset. The classifications are 1) D vs. S vs. V, 2) P vs. S vs. V, 3) D vs. P vs. S, and 4) D vs. P vs. V. We achieve the highest accuracy of 95.57%, 94.52%, 94.81%, and 86.15% for classification number 1), 2), 3), and 4), respectively. In addition, in all four wound-class classifications, the highest accuracy is achieved with the multi-modal (WMC) networks. Table 7 shows the detailed results of these classifications.

**Table 7:** Four Three-Wound-Class Classifications on AZH Dataset

| Classifications | | D-S-V | P-S-V | D-P-S | D-P-V |
|---|---|---|---|---|---|
| **Input** | **Model** | **Accuracy** | **Accuracy** | **Accuracy** | **Accuracy** |
| Location | MLP | 80.67% | 78.26% | 65.57% | 76.76% |
| | LSTM | 75.33% | 69.56% | 67.21% | 79.58% |
| Image | VGG16 | 74.67% | 68.12% | 61.48% | 76.06% |
| | VGG19 | 76.00% | 70.23% | 58.20% | 68.31% |
| Image + Location | VGG16+MLP | 86.67% | **86.23%** | 68.03% | 80.28% |
| | VGG19+MLP | **90.67%** | 81.16% | 72.13% | **80.99%** |
| | VGG16+LSTM | 85.33% | 75.36% | **72.95%** | 74.65% |
| | VGG19+LSTM | 87.33% | 68.16% | 67.21% | 70.42% |

Finally, ten binary classifications are performed on the AZH dataset. The classifications are: 1) N vs. D, 2) N vs. P, 3) N vs. S, 4) N vs. V, 5) D vs. P, 6) D vs. S, 7) D vs. V, 8) P vs. S, 9) P vs. V, and 10) S vs. V. We achieve highest accuracy of 100%, 98.31%, 98.51%, 100%, 85.00%, 88.64%, 92.59%, 81.58%, 89.58%, and 98.08% for classification number 1), 2), 3), 4), 5), 6), 7), 8), 9), and 10), respectively. In all binary classifications, the highest accuracy is achieved with the multi-modal (WMC) networks. Table 8 shows the detailed results of these binary classifications. The precision, recall, and f1-score for all the best models (according to accuracy) are also calculated and shown in Table 9.

**Table 8:** Accuracy of Ten Binary Classifications on AZH Dataset

| Classifications | | N-D | N-P | N-S | N-V | D-P | D-S | D-V | P-S | P-V | S-V |
|---|---|---|---|---|---|---|---|---|---|---|---|
| **Input** | **Model** | **Accuracy** | | | | | | | | | |
| Location | MLP | 76.06% | 64.04% | 74.63% | 77.01% | 73.75% | 75.00% | 87.96% | 73.68% | 83.33% | 90.38% |
| | LSTM | 78.87% | 42.37% | 74.63% | 78.16% | 75.00% | 82.95% | 57.41% | 72.37% | 83.33% | 93.27% |
| Image | VGG16 | 98.59% | 96.61% | 97.01% | 98.85% | 81.25% | 79.55% | 87.96% | 77.63% | 84.38% | 84.62% |
| | VGG19 | 98.59% | **98.31%** | 97.01% | 98.85% | 71.25% | 80.68% | 87.96% | 73.68% | 86.46% | 86.54% |
| Image + Location | VGG16 +MLP | 95.77% | 96.61% | **98.51%** | 97.70% | **85.00%** | **88.64%** | 90.74% | **81.58%** | **89.58%** | 93.27% |
| | VGG19 +MLP | 95.77% | 96.61% | 97.01% | 98.85% | 80.00% | 82.85% | **92.59%** | 77.63% | **89.58%** | **98.08%** |
| | VGG16 +LSTM | 97.18% | 93.22% | 97.01% | 98.85% | 80.00% | 78.41% | 91.67% | 75.00% | 86.46% | 86.54% |
| | VGG19 +LSTM | **100%** | **98.31%** | 97.01% | **100%** | 80.00% | 79.55% | 87.96% | 76.32% | 84.38% | 79.81% |

**Table 9:** Precision, Recall, F1-Score of Ten Binary Classifications on AZH Dataset

| Classifications | Best Model(s) | Precision | Recall | F1-Score |
|---|---|---|---|---|
| **N-D** | VGG19+LSTM | 100% | 100% | 100% |
| **N-P** | VGG19 | 100% | 97.06% | 98.51% |
| | VGG19+LSTM | 100% | 97.06% | 98.51% |
| **N-S** | VGG16+MLP | 100% | 97.62% | 98.80% |
| **N-V** | VGG19+LSTM | 100% | 100% | 100% |
| **D-P** | VGG16+MLP | 75.00% | 97.06% | 84.62% |
| **D-S** | VGG16+MLP | 83.33% | 95.24% | 88.89% |
| **D-V** | VGG19+MLP | 89.71% | 98.39% | 93.85% |
| **P-S** | VGG16+MLP | 75.00% | 100% | 85.71% |
| **P-V** | VGG16+MLP | 87.14% | 98.39% | 92.42% |
| | VGG19+MLP | 87.14% | 98.39% | 92.42% |
| **S-V** | VGG19+MLP | 96.88% | 100% | 98.41% |

## B.3 Experiment on Medetec Dataset

A classification between all the classes is performed on the Medetec dataset. Table 10 shows the results of this three-wound-class classification (D vs. P vs. A+V). We achieve the highest accuracy of 86.67% with the multi-modal (WMC) network using the VGG19+LSTM combination, where the highest accuracy achieved from WLC and WIC networks are 81.11% and 82.22% using both MLP and LSTM, and VGG16 networks, respectively.

**Table 10:** Three-Wound-Class Classification (D vs. P vs. A+V) on Medetec Dataset

| Input | Model | Accuracy |
|---|---|---|
| Location | MLP | 81.11% |
| | LSTM | 81.11% |
| Image | VGG16 | 82.22% |
| | VGG19 | 77.78% |
| | InceptionV3 | 48.89% |
| Image + Location | VGG16+MLP | 83.33% |
| | VGG19+MLP | 85.56% |
| | InceptionV3+MLP | 77.78% |
| | VGG16+LSTM | 85.56% |
| | VGG19+LSTM | **86.67%** |
| | InceptionV3+LSTM | 74.44% |

## B.4 Experiment on AZHMT Dataset

A classification between all the classes is performed on the AZHMT dataset. Table 11 shows the results of this six-class classification (BG vs. N vs. D vs. P vs. S vs. A+V). We achieve the highest accuracy of 80.56% with the multi-modal (WMC) network using the VGG19+MLP combination. The highest accuracy achieved from WLC and WIC networks is 69.44% and 72.22% using MLP and VGG19 networks, respectively.

**Table 11:** Six-Class Classification (BG vs. N vs. D vs. P vs. S vs. A+V) on AZHMT Dataset

| Input | Model | Accuracy |
|---|---|---|
| Location | MLP | 69.44% |
| | LSTM | 69.14% |
| Image | VGG16 | 67.59% |
| | VGG19 | 72.22% |
| | InceptionV3 | 53.40% |
| Image + Location | VGG16+MLP | 75.31% |
| | VGG19+MLP | **80.56%** |
| | InceptionV3+MLP | 68.83% |
| | VGG16+LSTM | 70.99% |
| | VGG19+LSTM | 68.52% |
| | InceptionV3+LSTM | 70.06% |

A four-wound-class classification is performed on the AZHMT dataset. The classification is done among D, P, S, and A+V classes. We achieved the highest accuracy of 80.29% with the multi-modal (WMC) network using the VGG19+MLP combination. The highest accuracy achieved from WLC and WIC networks is 77.01% and 68.61% using LSTM and VGG16 networks, respectively. Table 12 shows the detailed results of this four-wound-class classification.

**Table 12:** Four-Wound-Class Classification (D vs. P vs. S vs. A+V) on AZHMT Dataset

| Input | Model | Accuracy |
|---|---|---|
| Location | MLP | 75.55% |
| | LSTM | 77.01% |
| Image | VGG16 | 68.61% |
| | VGG19 | 63.14% |
| | InceptionV3 | 50.00% |
| Image + Location | VGG16+MLP | 78.83% |
| | VGG19+MLP | **80.29%** |
| | InceptionV3+MLP | 72.26% |
| | VGG16+LSTM | 70.44% |
| | VGG19+LSTM | 65.33% |
| | InceptionV3+LSTM | 69.71% |

Finally, ten binary classifications are performed on the AZHMT dataset. The classifications are: 1) N vs. D, 2) N vs. P, 3) N vs. S, 4) N vs. A+V, 5) D vs. P, 6) D vs. S, 7) D vs. A+V, 8) P vs. S, 9) P vs. A+V, and 10) S vs. A+V. We achieve highest accuracy of 100%, 100%, 98.51%, 100%, 86.90%, 86.90%, 92.76%, 89.34%, 93.41%, and 93.80% for classification number 1), 2), 3), 4), 5), 6), 7), 8), 9), and 10), respectively. In all binary classifications, the highest accuracy is achieved with the multi-modal (WMC) networks. Table 13 shows the detailed results of these binary classifications on the AZHMT dataset. The precision, recall, and f1-score for all the best models (according to accuracy) are also calculated and shown in Table 14.

**Table 13:** Accuracy of Ten Binary Classifications on AZHMT Dataset

| Classifications | | N-D | N-P | N-S | N-A+V | D-P | D-S | D-A+V | P-S | P-A+V | S-A+V |
|---|---|---|---|---|---|---|---|---|---|---|---|
| **Input** | **Model** | **Accuracy** | | | | | | | | | |
| Location | MLP | 75.56% | 81.91% | 74.63% | 81.25% | 85.52% | 84.14% | 87.50% | 86.07% | 92.22% | 90.70% |
| | LSTM | 77.78% | 85.71% | 74.63% | 81.25% | 82.76% | 84.11% | 57.24% | 83.61% | 91.02% | 90.70% |
| Image | VGG16 | 97.78% | 97.14% | 97.01% | 99.11% | 77.93% | 76.55% | 85.53% | 74.59% | 77.25% | 86.82% |
| | VGG19 | 97.78% | 98.10% | 97.01% | **100%** | 80.69% | 77.93% | 86.84% | 75.41% | 83.23% | 81.40% |
| Image + Location | VGG16 +MLP | 97.78% | 99.05% | **98.51%** | 98.21% | **86.90%** | **86.90%** | **92.76%** | 88.52% | 92.82% | 90.70% |
| | VGG19 +MLP | **100%** | 98.10% | 97.01% | 99.11% | 85.52% | 82.76% | 92.11% | 84.43% | **93.41%** | **93.80%** |
| | VGG16 +LSTM | 97.78% | 98.10% | 97.01% | 99.11% | 82.07% | 82.76% | 85.53% | **89.34%** | 87.42% | 83.72% |
| | VGG19 +LSTM | **100%** | **100%** | 97.01% | **100%** | 79.31% | 71.72% | 86.84% | 78.69% | 82.63% | 85.27% |

**Table 14:** Precision, Recall, F1-Score of Ten Binary Classifications on AZHMT Dataset

| Classifications | Best Model(s) | Precision | Recall | F1-Score |
|---|---|---|---|---|
| **N-D** | VGG19+MLP | 100% | 100% | 100% |
| | VGG19+LSTM | 100% | 100% | 100% |
| **N-P** | VGG19+LSTM | 100% | 100% | 100% |
| **N-S** | VGG16+MLP | 100% | 97.62% | 98.80% |
| **N-A+V** | VGG19 | 100% | 100% | 100% |
| | VGG19+LSTM | 100% | 100% | 100% |
| **D-P** | VGG16+MLP | 83.52% | 95.00% | 88.89% |
| **D-S** | VGG16+MLP | 81.44% | 98.75% | 89.27% |
| **D-A+V** | VGG16+MLP | 91.30% | 96.55% | 93.85% |
| **P-S** | VGG16+LSTM | 82.22% | 88.10% | 85.06% |
| **P-A+V** | VGG19+MLP | 89.58% | 98.85% | 93.99% |
| **S-A+V** | VGG19+MLP | 94.38% | 96.55% | 95.45% |

## C. Result Comparison with Previous Works

Classification results depend on many factors like dataset, model, training-validation-testing split, balanced or unbalanced dataset, resources used for training, etc. Though the datasets and other factors between our work and previous classification works are not the same, this section mainly focuses on how the multimodality using both image and location data can improve the classification accuracy. The comparison with the previous works is only made if all the classes of that work's dataset are present in our dataset. Our previous work [16]'s dataset is most similar to the work presented in this manuscript. Alongside [16], the classifications performed in [12], [13], and [15] have the classes that are present in our dataset. A detailed comparison between previous works and our current work is shown in Table 15.

The reasons why other related works are not considered in this comparison are: [10] performs burn vs. pressure ulcer classification, and our datasets do not contain any burn images; [11] performs binary classification of ischemia vs. non-ischemia and infection vs. non-infection on DFU images, which is not compatible with our datasets; [14] performs binary classifications between such kind of wounds (wound, infection (SSI), granulation tissue, etc.), which are not present in our datasets; and [17] performs multi-class wound classifications among diabetic, lymphovascular, pressure injury, and surgical wounds and our datasets do not contain the lymphovascular wound type.

**Table 15:** Comparison Among the Previous Works and the Present Work.

| Work | Classification | Evaluation Metrics | Previous Work | | | Present Work | | |
|---|---|---|---|---|---|---|---|---|
| | | | Model | Dataset | Result | Model | Dataset | Result |
| Goyal et al. [12] | Healthy Skin Vs. DFU Skin (N vs. D) | Accuracy | DFUNet | A dataset containing 397 wound images | 92.5% | VGG19 +LSTM | AZH | **100%** |
| | | | | | | VGG19 +MLP & VGG19 +LSTM | AZHMT | **100%** |
| Aguirre et al. [13] | VLU versus non-VLU (N vs. V, D vs. V, P vs. V, S vs. V) | Accuracy | VGG19 | A dataset of 300 wound images | 85% | **N-V:** VGG19 +LSTM | AZH | **100%** |
| | | | | | | **D-V:** VGG19 +MLP | AZH | **92.59%** |
| | | | | | | **P-V:** VGG19 +MLP | AZH | **89.58%** |
| | | | | | | **S-V:** VGG19 +MLP | AZH | **98.08%** |
| Alzubaidi et al. [15] | Normal Skin Vs. Abnormal (DFU) Skin (N vs. D) | F1-Score | DFU_QUT Net + SVM | A dataset containing 754-foot images | 94.5% | VGG19 +LSTM | AZH | **100%** |
| | | | | | | VGG19 +MLP & VGG19 +LSTM | AZHMT | **100%** |
| Rostami et al. [16] | S-V | Accuracy | An end-to-end Ensemble DCNN-based Classifier | A new dataset containing 538 wound images | 96.4% | VGG19 +MLP | AZH | **98.08%** |
| | D-S-V | " | " | " | **91.9%** | VGG19+MLP | " | 90.67% |
| | BG-N-D-V | " | " | " | 89.41% | VGG16+MLP | " | **95.57%** |
| | BG-N-P-V | " | " | " | 86.57% | VGG16 +LSTM | " | **94.52%** |
| | BG-N-S-V | " | " | " | 92.20% | VGG16+MLP | " | **94.81%** |
| | BG-N-D-P | " | " | " | 80.29% | VGG19+MLP | " | **86.15%** |
| | BG-N-D-S | " | " | " | **90.98%** | VGG19+MLP | " | 89.86% |
| | BG-N-P-S | " | " | " | 84.12% | VGG19 +LSTM | " | **88.89%** |
| | BG-N-D-P-V | " | " | " | 79.76% | VGG19+MLP | " | **84.38%** |
| | BG-N-D-S-V | " | " | " | 84.94% | VGG19+MLP | " | **90.00%** |
| | BG-N-D-P-S | " | " | " | **81.49%** | VGG16+MLP | " | 77.33% |
| | BG-N-P-S-V | " | " | " | **83.53%** | VGG16 +LSTM | " | 80.32% |
| | BG-N-D-P-S-V | " | " | " | 68.69% | VGG19+MLP | " | **82.91%** |

### D. Discussion

From all the result comparisons, we can see that using the multi-modal network with wound image and location improves the classification performance by a significant margin. However, before going into deeper analysis, it should be clear that there are two types of classifications: 1) mixed-class classifications (e.g., three-class classification, five-class classification, etc.), and 2) wound-class classifications (e.g., four wound-class classification, three wound-class classifications, etc.). The wound-class classification does not contain any non-wound classes (i.e., normal skin and background), and they are more challenging to classify than the mixed-class classification.

On the AZH dataset, for mixed-class classifications, we perform one six-class, four five-class, six four-class, and four binary classifications; and for wound-class classifications, we perform one four-wound-class, four three-wound-class, and six binary classifications. For the six-class classification, the WLC classifier performs the worst as there are some overlaps among locations of different classes, and the WMC classifier performs the best. From table 4, the best performance of WLC is achieved by MLP (60.68%), WIC is achieved by VGG16 (75.64%), and WMC is conducted by VGG19+MLP (82.91%). From table 5, for BGNDPV, BGNDSV, BGNDPS, and BGNPSV classifications, the best performance of WLC is achieved by LSTM, MLP, LSTM, and LSTM; the best performance of WIC is achieved by VGG19, VGG19, VGG19, and VGG16; and best performance of WMC is achieved by VGG19+MLP, VGG19+MLP, VGG16+MLP, and VGG16+LSTM, respectively. From table 6, we can also see the same consistency of the model performances. Though a single model of WLC or WIC, or a single combination of WMC does not always produce the best performance, the WMC classifier always performs the best in comparison to the WIC or WLC classifiers. The performance comparison of mixed-class classifications among the best models from each category (location, image, and multimodality) is shown in Figure 5. The same pattern can also be seen in the wound class classifications. The performance comparison among the best models of wound-class classifications from each category (location, image, and multimodality) is shown in Figure 6.

From Figure 5, the lowest accuracy is produced by BGNDPS (77.33%), and from Figure 6, the most insufficient accuracy is produced by DPS (72.95%). So, separating diabetic, pressure, and surgical wound is the hardest, according to our experiments. In the DPS classification, the highest accuracies achieve by WLC and WIC networks are 67.21% and 61.48%, respectively, which is much lower than any other classification results. So, distinguishing among these three (D, P, and S) classes is very hard with image or location data separately (single-modality). Also, from Figure 6, among all binary classifications, D vs. P and P vs. S classifications have the lowest accuracy of 85% and 81.58%, respectively. So, we can say that differentiation between pressure and surgical wounds is the most complicated task. From Figure 5, the highest accuracy is achieved by ND, NP, NS, and NV classifications with 100%, 98.31%, 98.51%, and 100%, respectively. Also, from Figure 6, the highest accuracy is achieved by SV classification with 98.08% accuracy. So, differentiating between normal skin and other wound types (D, V, S, and P) and differentiating between surgical wounds and venous leg ulcers are the most straightforward classifications task for our developed WMC classifier. Finally, from Figures 5 and 6, we can see that multimodality using wound image and location (WMC) performs best in comparison with single (image or location) modality (WLC or WIC) in all scenarios on the AZH dataset. Also, mixed-class classification results are comparatively higher than wound-class classifications.

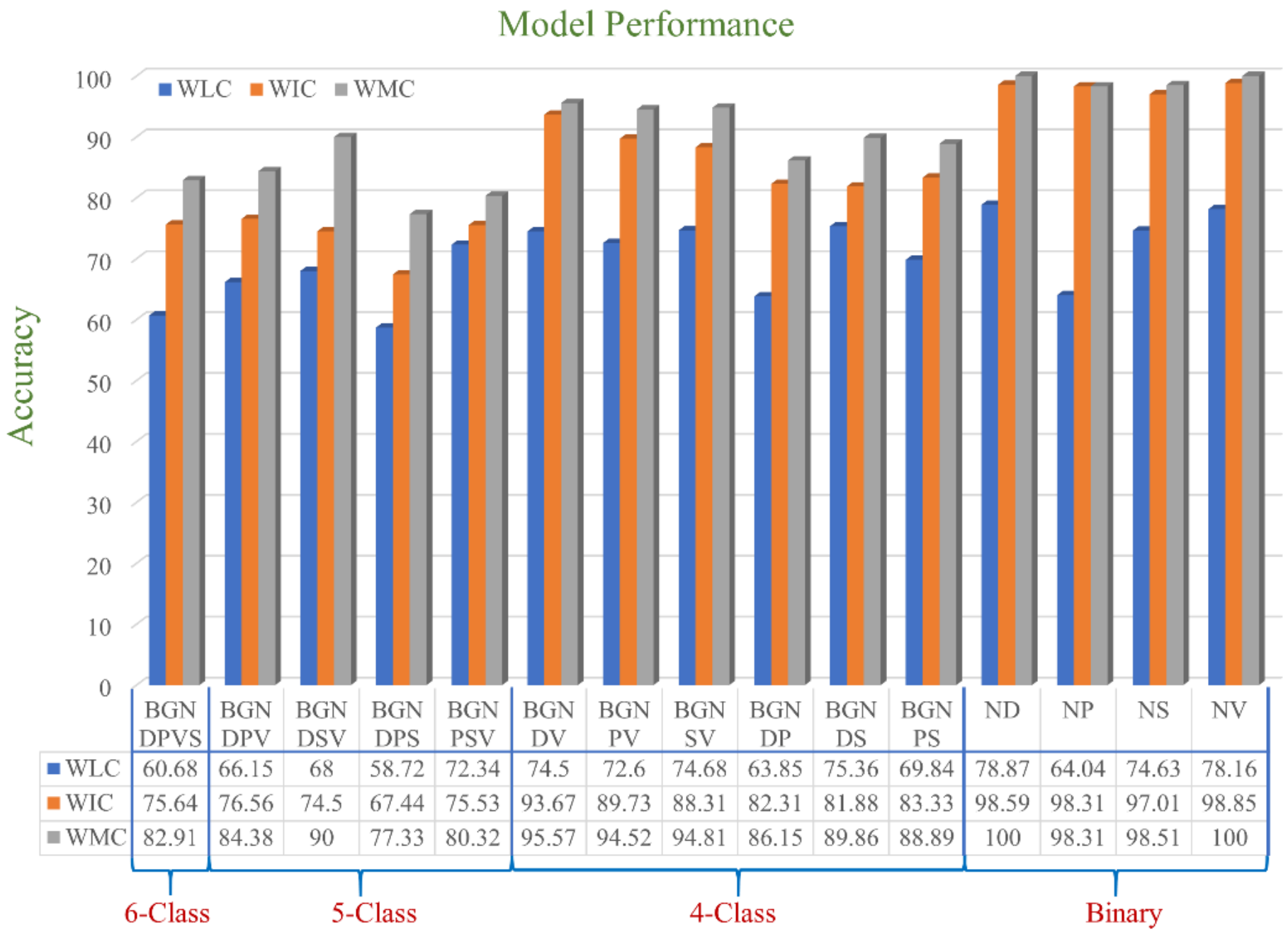

**Figure 5:** Performance comparison of mixed-class classification among the best models from each category (location -WLC, image-WIC, and multimodality-WMC) on AZH Dataset.

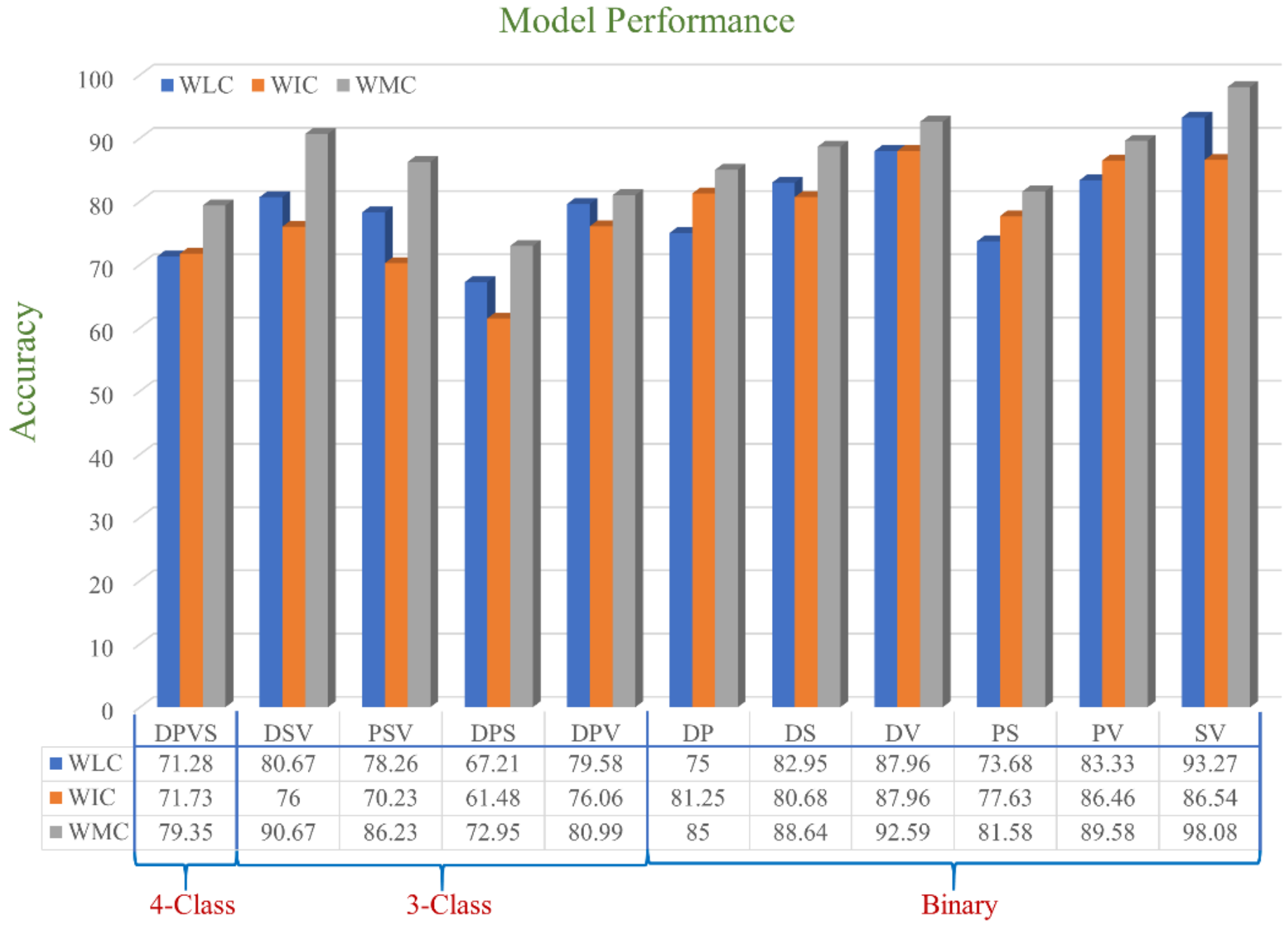

**Figure 6:** Performance comparison of wound-class classification among the best models from each category (location -WLC, image-WIC, and multimodality-WMC) on AZH Dataset.

On the Medetec dataset, we perform only one wound-type classification among all three classes (D, P, and A+V). Here, the highest accuracy is achieved by the WMC classifier. The highest accuracies achieve by WLC, WIC, and WMC classifiers are 81.11%, 82.22%, and 86.67%, respectively. Both MLP and LSTM perform the same for the classification among wound location data, VGG16 performs best for wound image classification, and VGG19+LSTM combination works best for this classification task.

On the AZHMT dataset, for mixed-class classification, we perform one six-class classification (BG-N-D-P-S-A+V) and four binary classifications (N-D, N-P, N-S, and N-A+V). We also perform one four-wound-class classification (D-P-S-A+V) and six wound-class binary classifications (D-P, D-S, D-A+V, P-S, P-A+V, and S-A+V). The performance comparison of mixed-class classifications among the best models from each category (location, image, and multimodality) on the AZHMT dataset is shown in Figure 7. The performance comparison among the best models of wound-class classifications from each category on the AZHMT dataset is shown in Figure 8.

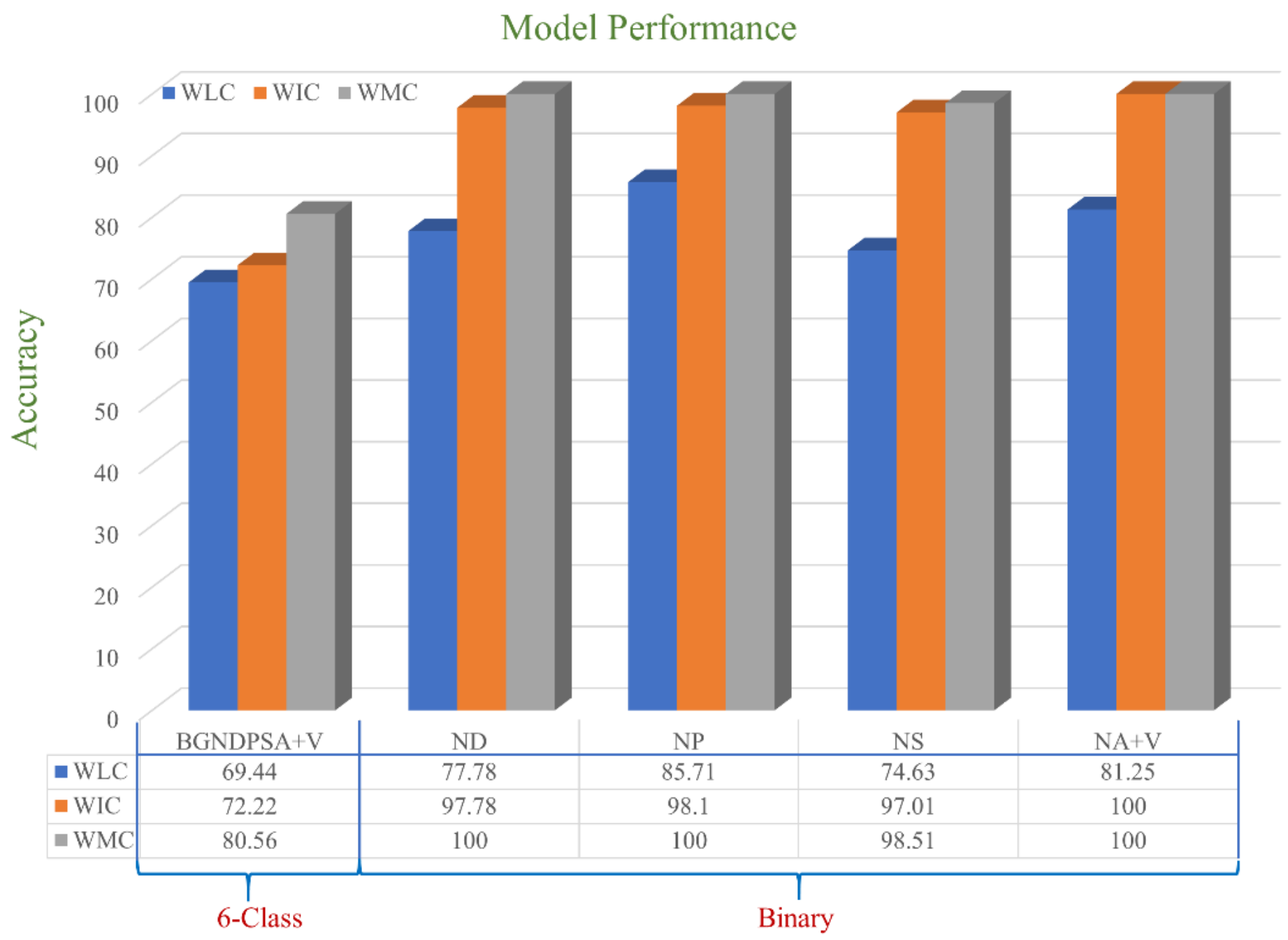

**Figure 7:** Performance comparison of mixed-class classification among the best models from each category (location -WLC, image-WIC, and multimodality-WMC) on AZHMT Dataset.

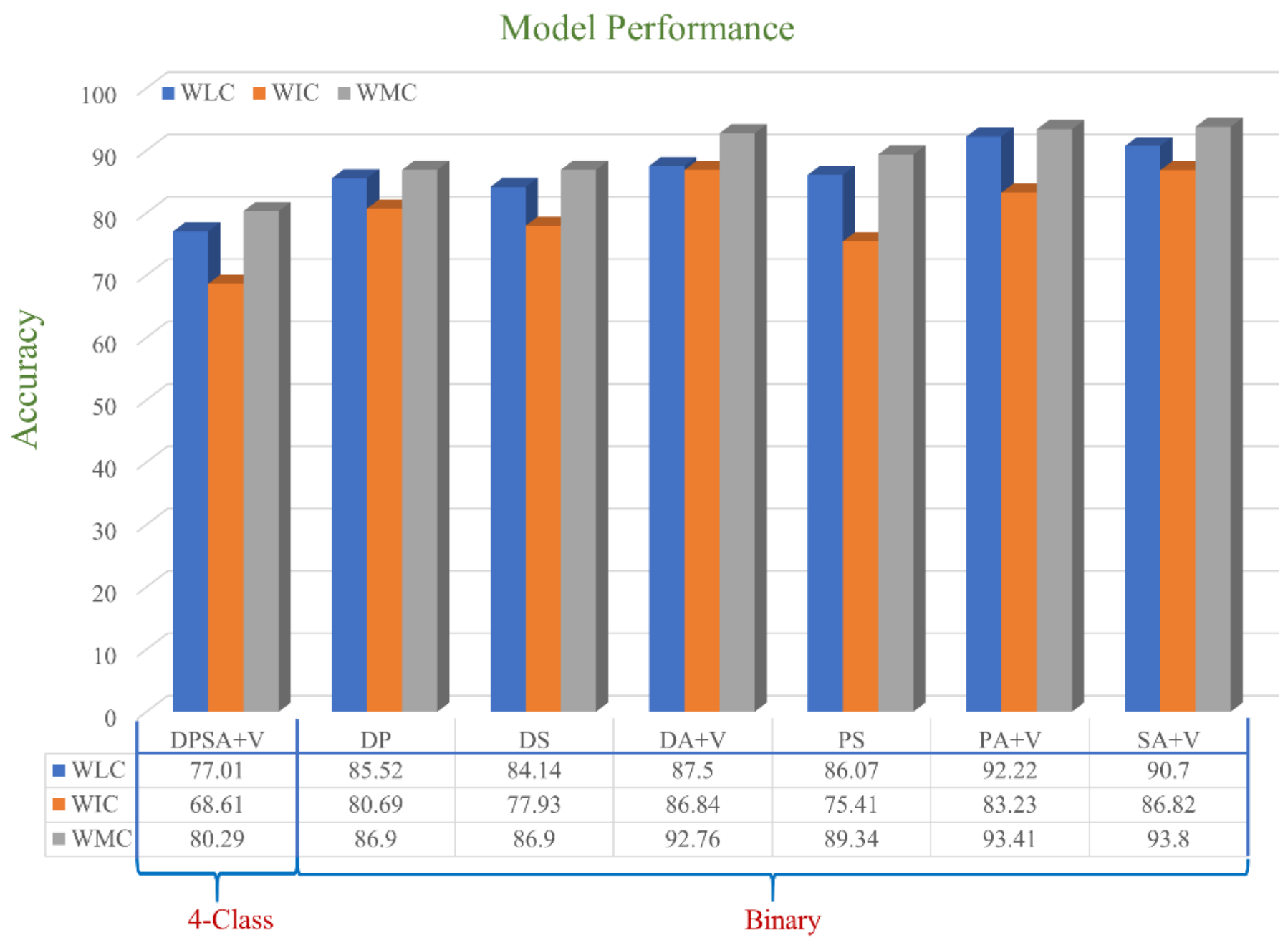

**Figure 8:** Performance comparison of wound-class classification among the best models from each category (location -WLC, image-WIC, and multimodality-WMC) on AZHMT Dataset.

From Figure 7, the lowest accuracy is achieved by the six-class classification, as it is more complex than the binary classifications. Among four binary classifications, three (N-D, N-P, and N-A+V) attain an accuracy of 100%. In all these classifications, the WMC network outperforms WLC and WIC networks. From Figure 8, the four-wound-class classification achieves the highest accuracy of 80.29% using the VGG19+MLP combination of WMC network, and the WIC network achieves the lowest accuracy (68.61%). Among the six binary classifications in Figure 8, DP and DS have the combined lowest accuracy of 86.9%, followed by PS (89.4%). As discussed earlier, according to these results separating diabetic, pressure, and surgical wound is the hardest according to our experiments. Here the highest accuracy is achieved by SA+V classification with 93.80% accuracy. So, we come to the same conclusion as discussed in AZH dataset experiments that differentiation between normal skin and other wound types (D, A+V, S, and P) and differentiation between surgical wounds and arterial ulcers and venous leg ulcers are the most straightforward classifications task for our developed WMC classifier.

In Figures 5 and 7, the WLC network performance is abysmal compared to the WIC and WMC network. One important reason is that there are some overlaps among the normal (healthy) skin and other wound classes, as the normal skin is cropped from the wound images. In one patient's wound image, a non-infected (normal) skin can be infected in another patient's wound image, which produces these overlaps and thus decreases the WLC performance. Figures 6 and 8 show that the WLC network's performance is better than the WIC network as there is no normal skin (N) class in these classifications. But still, the WLC network performance can be improved by increasing the number of data, which can help increase the WMC network's performance in the long run. Figure 9 shows some examples of location overlapping among different classes.

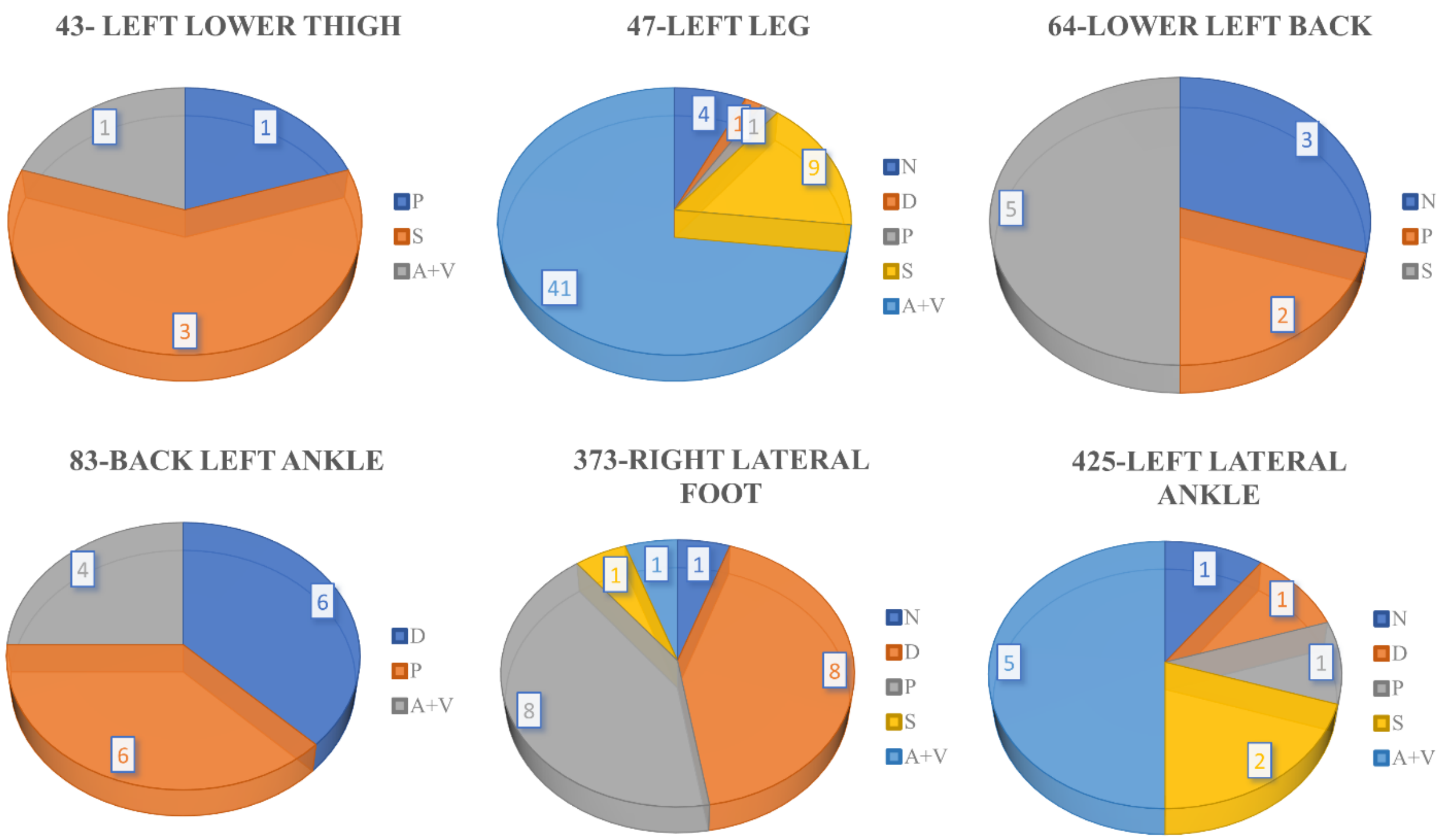

**Figure 9:** Examples of Location Overlaps on AZHMT dataset.

AZHMT is a mixed dataset that contains both wound image and location data from AZH and Medetec datasets. Comparing these results of AZH and AZHMT datasets, we can see that both perform very well with different classifications. A comparison between the highest results (accuracy) of AZH and AZHMT datasets is shown in Figure 10. As mentioned earlier, all the results are from the multi-modal network (WMC), as it outperforms all the single modal (WIC and WLC) networks. The results are very similar for some classifications for both datasets. Here, ND, NS, and NV/NA+V have the same highest accuracy. The N and S class contain the same images and locations in both datasets, as there is no surgical wound (S) class in Medetec dataset. For the six-class classification, the AZH dataset has 2.35% more accuracy than the AZHMT dataset. For the four-wound-class classification, the AZHMT dataset has 0.94% more accuracy than the AZH dataset. In the binary wound classifications, DP, DS, DV/DA+V, and PV/PA+V have very similar results for both datasets, where in some cases, AZH has slightly more accuracy than AZHMT and vice versa. For the PS classification, the AZHMT dataset has 7.76% more accuracy than the AZH dataset, and for the SV/SA+V classification, the AZH dataset has 4.28% more accuracy than the AZHMT dataset. Both datasets have some pros and cons that is reflected in these results: AZHMT contains more data than the AZH dataset, which is an advantage for training deep learning models; but AZHMT also contains mixed data from two sources, which makes the dataset more challenging to classify; AZHMT also contains a mixed data on a single class (arterial and venous ulcer combination), which may also impact the results. Overall, these results on both AZH and AZHMT datasets prove the reliability and robustness of our developed WMC network for wound classifications.

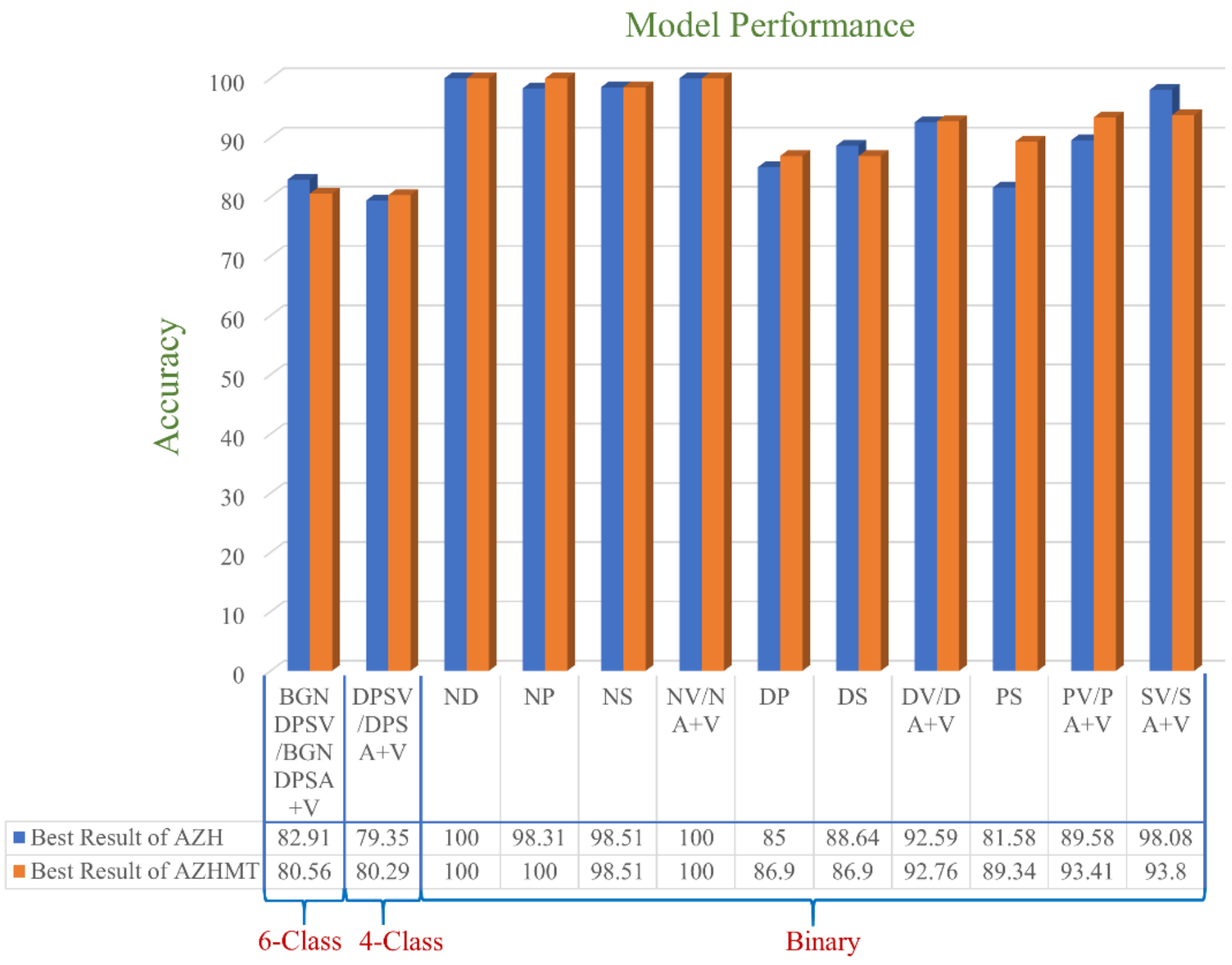

**Figure 10:** Comparison between the highest results (accuracy) of AZH and AZHMT datasets.

From Table 15, we can see that our work outperforms all the previous works by a huge margin, except for some exceptions with our previous work [16]. As mentioned earlier, this comparison is not perfect as factors like dataset, model, training-validation-testing split, balance ness of the dataset, resources used for training, etc., are not the same as the previous works. But this comparison proves that multimodality using wound image and location can improve the wound classification results by a huge margin. We achieve a 7.5% improvement of accuracy for classifying Healthy Skin Vs. DFU Skin (N Vs. D) from Goyal et al.'s work [12], with both of our AZH and AZHMT datasets. Comparing to Aguirre et al.'s work [13] of classifying VLU versus non-VLU (V vs. [N or D or P or S]) wounds, we achieve a significant 4.58% to 15% improvement of accuracy with the AZH dataset. In this experiment, we got an improvement of 4.58% for VLU vs. PU, 7.59% for VLU vs. DFU, 13.08% for VLU vs. Surgical, and 15% for VLU vs. Normal skin. Our developed classifier outperformed Alzubaidi et al.'s work [15] of Normal Skin Vs. Abnormal (DFU) Skin (N vs. D) classification with 5.5% improvement in F1-score for both the AZH and AZHMT experiments. Finally, compared to our previous work [16], there are 13 similar experiments in our present work. Among these 13 experiments, our current work got slightly lower accuracy in 4 experiments, and in the other 9 experiments, we achieved a significant improvement with the multi-modal WMC network. The 4 experiments where WMC has lower accuracy than our previous work are: 1) DSV classification with 1.23% difference, 2) BGNDS classification with 1.12% difference, 3) BGNDPS classification with 4.16% difference, and 4) BGNPSV classification with 3.21% difference. The 9 experiments where WMC has higher accuracy than our previous work are: 1) SV classification with 1.68% difference, 2) BGNDV classification with 6.16% difference, 3) BGNPV classification with 7.95% difference, 4) BGNSV classification with 2.61% difference, 5) BGNDP classification with 5.86% difference, 6) BGNPS classification with 4.77% difference, 7) BGNDPV classification with 4.62% difference, 8) BGNDSV classification with 5.06% difference, and 9) BGNDPSV classification with 14.22% difference. Both of works have some pros and cons: in the previous work, we have a balanced dataset (all classes had the same no of images), where the current work has an unbalanced dataset (Table 1); the previous work uses a very sophisticated ensemble classifier for image classification, where this work uses simple transfer learning with available DNN networks (VGG16, VGG19, etc.); the previous work only used wound images for training the classifier, where the current network uses both wound images and their corresponding locations for developing the classifier. Overall, we can say that this work outperforms nearly all the previous works.

## V. Conclusion

This paper developed a multi-modal wound classifier (WMC) network using wound images and their corresponding locations to classify wound into different classes. To the best of our knowledge, it is the first developed multi-modal network that uses images and locations for wound classification. This research is also the first work that classifies wounds according to their locations. To prepare the location data, we also developed a body map to help clinicians document the wound locations in the patient's record. The developed body map is currently used in the AZH wound center for location tagging to avoid inconsistency with location information. Three datasets with wound images and their corresponding locations are also developed and labeled by wound specialists of AZH wound center to perform many wound classification experiments. The multi-modal (WMC) network is created in the concatenation of two networks: wound image classifier (WIC) and wound location classifier (WMC). Developing the WIC network transfer learning is used with top-rated deep learning models. The WLC network is also developed by using deep learning models that are popular for controlling categorical data. A large number of experiments with a range of binary to six-class classifications are performed in three datasets, where many wound classifications are never performed before, to the best of our knowledge. The results produced by the WMC network are much better than the results produced from the WIC or WLC networks, and also these results beat nearly all the previous experimental results. In future experiments, the performance of the WMC network can be improved further by using more specific WIC and WLC networks for wound image classifications and wound location classifications, respectively. There are some overlaps in the wound location data, for which the WLC network produces lower accuracy comparatively to WIC and WMC networks. Increasing the number of data can improve the location (WLC) classifier. Overall, the developed WMC classifier can significantly speed up the automation of wound healing systems in the near future.